\documentclass{template/bmvc2k}
\usepackage{graphicx}
\usepackage{booktabs}
\usepackage{ulem}
\usepackage{amsfonts}
\usepackage{xcolor}
\usepackage[capitalise]{cleveref}

\newcommand{\boldstart}[1]{\vspace{4pt}\noindent\textbf{#1}}
\newcommand{\methodname}{\text{Ray-ONet}\xspace}

\def\ie{\textit{i.e.}, }
\usepackage[title]{appendix}
\usepackage{wrapfig}

\title{Ray-ONet: Efficient 3D Reconstruction \\ From A Single RGB Image}

\addauthor{Wenjing Bian}{wenjing@robots.ox.ac.uk}{1}
\addauthor{Zirui Wang}{ryan@robots.ox.ac.uk}{1}
\addauthor{Kejie Li}{kejie.li@eng.ox.ac.uk}{1}
\addauthor{Victor Adrian Prisacariu}{victor@robots.ox.ac.uk}{1}

\addinstitution{
 Active Vision Lab\\
 University of Oxford\\
 Oxford, UK
}

\runninghead{Bian et al.}{Efficient 3D Reconstruction From A Single RGB Image}
\def\eg{\emph{e.g}\bmvaOneDot}

\def\etal{\emph{et al}\bmvaOneDot}

\begin{document}

\maketitle

\vspace{-0.7cm}
\begin{figure}[h]
    \centering
    \includegraphics[width=0.85\textwidth]{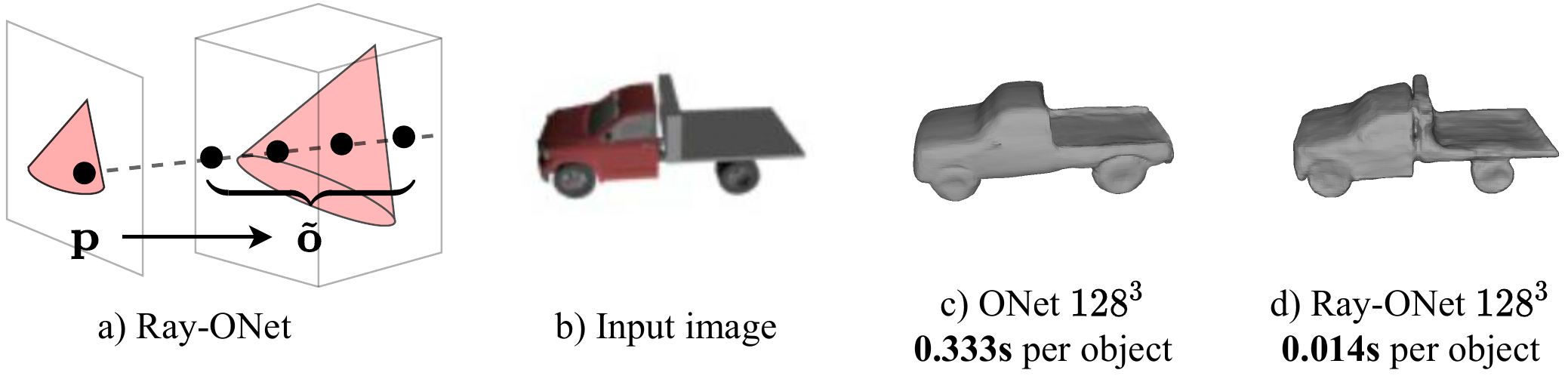}
    \caption{\methodname recovers more details than ONet~\cite{mescheder2019occupancy} while being $20\times$ faster, by predicting a series of occupancies $\tilde{\mathbf{o}}$ along the ray that is back-projected from a 2D point $\mathbf{p}$.}
    \vspace{-0.5cm}
    \label{fig:page1}
\end{figure}

\begin{abstract}
We propose \methodname to reconstruct detailed 3D models from monocular images efficiently. 
By predicting a series of occupancy probabilities along a ray that is back-projected from a pixel in the camera coordinate, our method \methodname improves the reconstruction accuracy in comparison with Occupancy Networks (ONet), while reducing the network inference complexity to O($N^2$).
As a result, \methodname achieves state-of-the-art performance on the ShapeNet benchmark with more than 20$\times$ speed-up at $128^3$ resolution and maintains a similar memory footprint during inference. The project website is available at \url{https://rayonet.active.vision}.
\end{abstract}
\vspace{-0.3cm}

\section{Introduction} \label{sec:intro}
3D reconstruction from a single view is an inherently ill-posed problem. However, it is trivial for a human to hallucinate a full 3D shape of a chair from a single image, since we have seen many different types of chairs from many angles in our daily life. 
Therefore, recovering a plausible 3D geometry from a single view is possible when enough prior knowledge is available, and in certain situations, might be the only feasible method when multi-view constraints are inaccessible, for instance, creating a 3D chair model from a chair image on a random website for an Augmented Reality (AR) application.

Recently, the single-view object reconstruction field has been advanced significantly using deep learning. In particular, deep implicit representations, such as DeepSDF~\cite{park2019deepsdf}, Occupancy Networks~\cite{mescheder2019occupancy} and their extensions~\cite{ saito2019pifu, peng2020convolutional}, demonstrate promising results on the single-view 3D reconstruction task, by representing 3D shapes via a continuous function that is embedded in network parameters. 
Although these methods enjoy the advantages of representing 3D models in any resolutions and require small memory footprint and low storage space, the inference speed is still an issue yet to address, as most existing methods require querying millions of 3D locations to recover one 3D shape, leading to an O$(N^3)$ complexity when a simple 3D sampling strategy is employed.


In this paper, we propose to predict a series of occupancies over 3D positions along a ray back-projected from a query 2D location on an image plane given local and global image features jointly. 
This formulation -- viewer-centred ray-based occupancy prediction, taking advantage of local and global image features -- improves the reconstruction accuracy while reducing the complexity to O$(N^2)$.

One issue of this viewer-centred reconstruction is the depth/scale ambiguity presented by a single view (\ie the same image can be rendered by a small object that is close to the camera or a large object that is at a distance). 
A common practice is to normalise all ground-truth shapes into a constant scale during training. 
However, this operation implicitly entangles the shape learning with the scale ambiguity.
Instead, we show how one can decouple these two problems and improve the reconstruction accuracy by providing the network with a scale calibration factor, which consists of a camera focal length and a camera-object distance.

In summary, we introduce \methodname with three contributions:
\emph{First}, our method predicts a series of occupancies over 3D positions along a ray from a 2D query point, by conditioning an occupancy prediction function to both global and local image context, reducing the complexity to O$(N^2)$.
\emph{Second}, we introduce a scale calibration factor, to compensate for the issue caused by different object scales.
\emph{Third}, we demonstrate on ShapeNet that our method improves the overall 3D reconstruction accuracy while being 20$\times$ faster than ONet.
\section{Related Work}
\label{sec:related}
\boldstart{Shape Representation.} 
Computation efficiency and reconstruction quality are both effected by the choice of representation for the model output. 
Researchers resort to volumetric representations~\cite{choy20163d, xie2019pix2vox} when deep networks are used for shape generation because 3D voxel grids naturally fit into the convolutional operation. 
However, the resolution of voxels is often limited by memory, which grows exponentially for high-resolution outputs. 
Alternatively, point clouds have been adopted~\cite{fan2017point}, but they cannot represent surfaces directly. 
Meshes~\cite{wang2018pixel2mesh}, being compact and surface aware, assume a pre-defined topology.

In recent years, implicit functions~\cite{mescheder2019occupancy,park2019deepsdf,chen2019learning} are widely used for representing continuous 3D shapes.
Networks using these implicit representations are trained such that the decision boundary represents the iso-surface of an object shape.
These models are able to reconstruct high-resolution 3D shapes with low memory cost, but the inference time increases cubically with sampling resolution.
Hybrid models~\cite{runz2020frodo, poursaeed2020coupling} which combine two representations have also been developed recently. 
They couple explicit and implicit representations with 2 branches, where the explicit branch accelerates surface reconstruction while the implicit branch reconstructs fine details of object shapes, but the network size is inevitably increased dramatically.
In contrast, \methodname is able to speed up inference without increasing the memory footprint by predicting occupancy probability along a ray in a single network forward.

\boldstart{Single-View 3D Reconstruction.} 
Reconstructing from a single view is a severely ill-posed problem, which was traditionally solved by ``shape-from-X'' techniques~\cite{zhang1999shape, laurentini1994visual} (X could be silhouette, shading etc.) using geometric or photometric information.
The resurgence of deep learning in recognition tasks has encouraged researchers to solve single-view 3D reconstruction in a data-driven manner.
PSGN~\cite{fan2017point} and 3D-R2N2~\cite{choy20163d} pioneer using a deep network to learn a mapping from an RGB image to an object shape represented by voxels or point clouds.
Following works are able to achieve better performance by using more sophisticated representations (\eg Octree~\cite{tatarchenko2017octree}, dense point clouds~\cite{lin2018learning}, or implicit representations~\cite{mescheder2019occupancy}) or building shape prior~\cite{wu2018learning}.

More recently, CoReNet~\cite{popov2020corenet} predicts translation equivariant voxel grid using a variable grid offset. 
It enables fast mesh extraction but still requires significantly more memory than implicit models. 
The closest to our work is SeqXY2SeqZ~\cite{han2020seqxy2seqz}, which also tries to reduce the time complexity of inference in implicit representations. To this end, they design a hybrid representation, which {interpret a 3D voxel grid as a set of tubes along a certain axis.} However, its reconstruction resolution is fixed in all 3 dimensions, because of the built-in location embedding matrices.

Our work is also a hybrid approach, which explicitly predicts occupancy probabilities along a certain ray that passes through a point on the input image. 
Unlike Han \etal~\cite{han2020seqxy2seqz}, we predict occupancies along a ray in a single network forward, which results in further reduction in time complexity.
In addition, the coordinate system of \methodname is aligned with the input image to take advantage of local features. As a result, we can predict 3D shapes with fine details.

\boldstart{Coordinate System.} 
Two types of coordinate systems are used in single-view object reconstruction: i) viewer-centred aligned to the input image or; ii) object-centred aligned to a canonical position in the shape database. 
As discussed in \cite{shin2018pixels, tatarchenko2019single}, an object-centred coordinate system with aligned objects of the same category encourages learning category information, which provides a robust prior for learnt categories, whereas predictions made in the view space rely more on geometric information from the input image, and can generalise better to unseen categories. 
Furthermore, recent works~\cite{popov2020corenet} using the viewer-centred coordinate system demonstrate reconstruction with fine details by using image local feature. However, they often assume 3D shapes are normalised to a constant scale.
Instead, we introduce a scale calibration factor to facilitate training given GT shapes in different scales.

\section{Preliminary} \label{sec:preliminary}
Occupancy network (ONet) \cite{mescheder2019occupancy} formulates shape reconstruction as a continuous function mapping problem, where the goal is to learn a function $\mathcal{F}$ that maps a 3D position $\mathbf{x} \in \mathbb{R}^{3}$ to an occupancy probability $\tilde{o}$, where $\{\tilde{o} \in \mathbb{R}: 0 < \tilde{o} < 1\}$.
The object shape is implicitly represented by the decision boundary of the learned function $\mathcal{F}$.
To capture shape variations, the occupancy function $\mathcal{F}$ above is often conditioned on observation, for example, an input image. Mathematically, the occupancy prediction function $\mathcal{F}$ can be defined as:
\begin{equation}
\label{eq:onet}
    \mathcal{F}(\mathbf{x}, \mathbf{z}) = \tilde{o},
\end{equation}
where $\mathbf{z}$ is a latent code that encodes an input image $I$.

This mapping function, however, suffers from slow inference speed because the network needs to query a large number of 3D positions to recover a high-resolution 3D reconstruction, which leads to O($N^3$) complexity if the 3D positions are sampled in a regular 3D grid.
Although more sophisticated sampling strategies such as Multi-resolution IsoSurface Extraction (MISE) can be applied as in ONet~\cite{mescheder2019occupancy}, the inference speed is still largely restrained by this query-every-3D-point strategy. 
This motivates us to develop a novel ray-based pipeline that reduces the complexity to O($N^2$).

\begin{figure}[]
    \centering
    \includegraphics[width=0.9\textwidth]{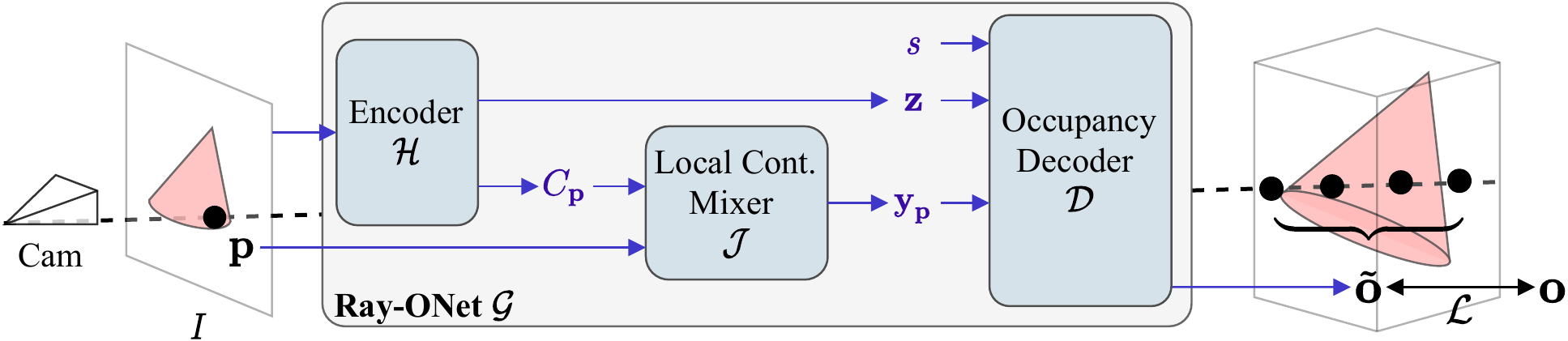}
    \caption{\textbf{\methodname Overview.} 
    Given an RGB image, \methodname first maps the image to a global feature descriptor $\mathbf{z}$ and a dense feature map $C$ via an image encoder $\mathcal{H}$. The local feature $C_\mathbf{p}$ (feature at position $\mathbf{p}$) and the position $\mathbf{p}$, taken by the local context mixer $\mathcal{J}$, are mapped to a local context descriptor $\mathbf{y_p}$. Finally, the occupancy decoder $\mathcal{D}$ takes as input $\mathbf{y_p}$, $\mathbf{z}$ and a scale factor $s$ and predicts occupancies of 3D points along the back-projected ray from position $\mathbf{p}$.
    }
    \vspace{-0.4cm}
    \label{fig:model}
\end{figure}

\section{Method}\label{sec:method}
Given an input image $I$, we aim to build a 3D reconstruction in the camera space of $I$, instead of in the canonical space as in ONet, by predicting series of occupancies along rays that are back-projected for all image pixels.

In other words, for each ray points to a 2D position $\mathbf{p}$ on the image plane, we predict a series of occupancies along the ray behind the 2D position in one forward pass, hence reducing the complexity to O($N^2$). 
As a ray direction can be uniquely defined by the 2D position $\mathbf{p}=(p_x, p_y)$ in the viewer space, our \methodname, which conditions the occupancy prediction on the ray direction, can be considered as conditioned on the 2D position $\mathbf{p}$. \cref{fig:model} presents an overview of our method.

In the following sections, we first formalise our occupancy prediction function in \cref{sec:method_occ_pred}, followed by how we assemble it with image encoders and fusion modules (\cref{sec:method_modules}) and a detail description of the scale calibration factor that facilitates training given various scales of ground-truth object shapes (\cref{sec:method_scale}). Lastly, we describe our training and inference pipeline in \cref{sec:method_train} and \cref{sec:method_inference}, including the training loss and the mesh extracting process etc.

\vspace{-0.25cm}
\subsection{Occupancy Prediction}\label{sec:method_occ_pred}
We now detail our occupancy prediction function. 
In \methodname, we propose a function $\mathcal{G}$:
\begin{equation}
    \mathcal{G}(\overbrace{\mathbf{p}, C_\mathbf{p}}^{\text{Local}}, \overbrace{\vphantom{C_\mathbf{p}}\mathbf{z}, s}^{\text{Global}}) = \tilde{\mathbf{o}},
    \label{eq:our_onet}
\end{equation}
which maps a 2D position $\mathbf{p}$ on an image $I$ to a number of $M$ occupancy estimations $\tilde{\mathbf{o}} \in \mathbb{R}^M$, conditioning on a global image feature $\mathbf{z}$, a local image feature $C_\mathbf{p}$ and a scale calibration factor $s$. 
Technically, the occupancy prediction function $\mathcal{G}$ is parameterised by a Multi-Layer Perceptron (MLP), and image global features $\mathbf{z}$ and local features $C_\mathbf{p}$ are encoded by a shared CNN encoder.
From now on, with a slight abuse of notation, \emph{we also refer $\mathcal{G}$ as our ray occupancy network}, as the network is just a parameterisation of the occupancy prediction function.

Comparing with the $\mathcal{F}(\mathbf{x}, \mathbf{z})$ in the classical ONet in \cref{eq:onet}, which maps a 3D position $\mathbf{x}$ to an occupancy estimation $\tilde{o}$ while conditioning on a global image feature, \methodname uses $\mathcal{G}(\mathbf{p}, C_\mathbf{p}, \mathbf{z}, s)$ to map a 2D position $\mathbf{p}$ to a series of occupancy estimations $\tilde{\mathbf{o}}$ by conditioning the occupancy function on both local and global context. \methodname, therefore, can predict all $M$ occupancies along a ray in a \emph{single network forward pass}, reducing the complexity to O$(N^2)$.
Moreover, as both our occupancy prediction function $\mathcal{G}$ and the classical $\mathcal{F}$ in ONet are parameterised in a similar MLP structure, our method maintains a similar memory footprint during inference.

\subsection{Individual Modules} \label{sec:method_modules}
Given an input image $I$ and its intrinsic and extrinsic camera parameters, we aim to extract relevant information for the occupancy prediction function $\mathcal{G}$. 
Specifically, we split the occupancy prediction function $\mathcal{G}$ to three modules: 
\emph{First}, an image encoder $\mathcal{H}$, which extracts image global feature $\mathbf{z}$ and local feature $C_\mathbf{p}$; 
\emph{Second}, a local context mixer $\mathcal{J}$, which fuses local image feature $C_\mathbf{p}$ with a ray that is defined by the 2D position $\mathbf{p}$ that it points to; 
and \emph{Third}, an occupancy decoder $\mathcal{D}$, which decodes a series of occupancy estimations from both local and global context.



\boldstart{Image Encoder $\mathcal{H}$.}
We extract a global feature vector $\mathbf{z}$ and a feature map $C$ from the input image $I$ using a ResNet-fashion CNN \cite{he2016resnet} $\mathcal{H}: I \rightarrow (\mathbf{z}, C)$.
Specifically, the global latent code $\mathbf{z}$ is an output of a ResNet-18, and the feature map $C \in \mathbb{R}^{H \times W \times D}$ is a result of concatenating bilinear upsampled activations after the ResNet blocks, where $H$ and $W$ denote input image resolution, and $D$ denotes the feature dimension.

\boldstart{Local Context Mixer $\mathcal{J}$.}
We parameterise the local context mixer module with an MLP $\mathcal{J}: (C_\mathbf{p}, \mathbf{p}) \rightarrow \mathbf{y}_\mathbf{p}$, which aggregates a 2D location $\mathbf{p}$ and an interpolated image feature $C_\mathbf{p}$ to a fused local context feature $\mathbf{y}_\mathbf{p}$.

\boldstart{Occupancy Decoder $\mathcal{D}$.}
For a local feature $\mathbf{y}_\mathbf{p}$ at point $\mathbf{p}$, we aim to predict a series of occupancies, conditioning on a global feature $\mathbf{z}$ and a scale calibration factor $s$, via a decoder $\mathcal{D}: (\mathbf{y}_\mathbf{p}, \mathbf{z}, s) \rightarrow \tilde{\mathbf{o}}$.
Similar to ONet, we parameterise the decoder $\mathcal{D}$ in an MLP network with skip links and Conditional Batch Normalisation (CBN).

To summarise this section, we assemble the occupancy prediction network $\mathcal{G}$ with three modules: an image encoder $\mathcal{H}$, a local context mixer $\mathcal{J}$ and an occupancy decoder $\mathcal{D}$:
\begin{equation}
    \mathcal{G}(\overbrace{\mathbf{p}, C_\mathbf{p}}^{\text{Local}}, \overbrace{\vphantom{C_\mathbf{p}}\mathbf{z}, s}^{\text{Global}})
    = \mathcal{D}(\mathcal{J}(\mathbf{p}, C_\mathbf{p}), \mathbf{z}, s)
    = \tilde{\mathbf{o}},
\end{equation}
where the global image feature $\mathbf{z}$ and the image feature map $C$ are extracted the input image $I$ from encoder $\mathcal{H}$.

\subsection{Scale Calibration} \label{sec:method_scale}
The scale of the predictions is as important as the reconstructed shape because a correct shape prediction but in a different scale to the ground truth can result in a large training loss. 
Instead of normalising all ground-truth shapes, we introduce the scale calibration factor $s$ as an extra dimension for the global feature to facilitate network training given ground-truth shapes in different scales.
Specifically, we define our scale calibration factor $s = \|\mathbf{c}\|_{2} / f$, where $\|\mathbf{c}\|_{2}$ denotes the distance between a camera and a ground truth 3D model, and $f$ denotes the camera focal length.

This scale calibration factor $s$ conveys that the predicted shape scale should be proportional to camera-object distance and inverse proportional to the camera focal length.
In the case of using a fixed focal length for the entire training set, the scale calibration factor can be further reduced to $\|\mathbf{c}\|_{2}$. 
Note that this scale calibration factor is only needed during training.

\subsection{Training}\label{sec:method_train}
We use a binary cross-entropy loss $\mathcal{L}$ between predicted occupancies $\tilde{\mathbf{o}}$ and ground truth occupancies $\mathbf{o}$ to supervise the training of the encoder $\mathcal{H}$, the mixer $\mathcal{J}$ and the decoder $\mathcal{D}$:
\begin{equation}
    \mathcal{L}(\tilde{\mathbf{o}}, \mathbf{o}) =
    -[\mathbf{o} \log(\tilde{\mathbf{o}}) 
    + (1 - \mathbf{o}) \log(1 - \tilde{\mathbf{o}})].
\end{equation}

As we predict 3D models in the camera space of an input image, we rotate ground truth 3D models so that the projection of a ground truth 3D model on the input image plane matches the observations on $I$. This step is necessary to acquire ground truth occupancies $\mathbf{o}$ in the camera space of an input image.

\begin{figure}
    \centering
    \includegraphics[width=0.95\textwidth]{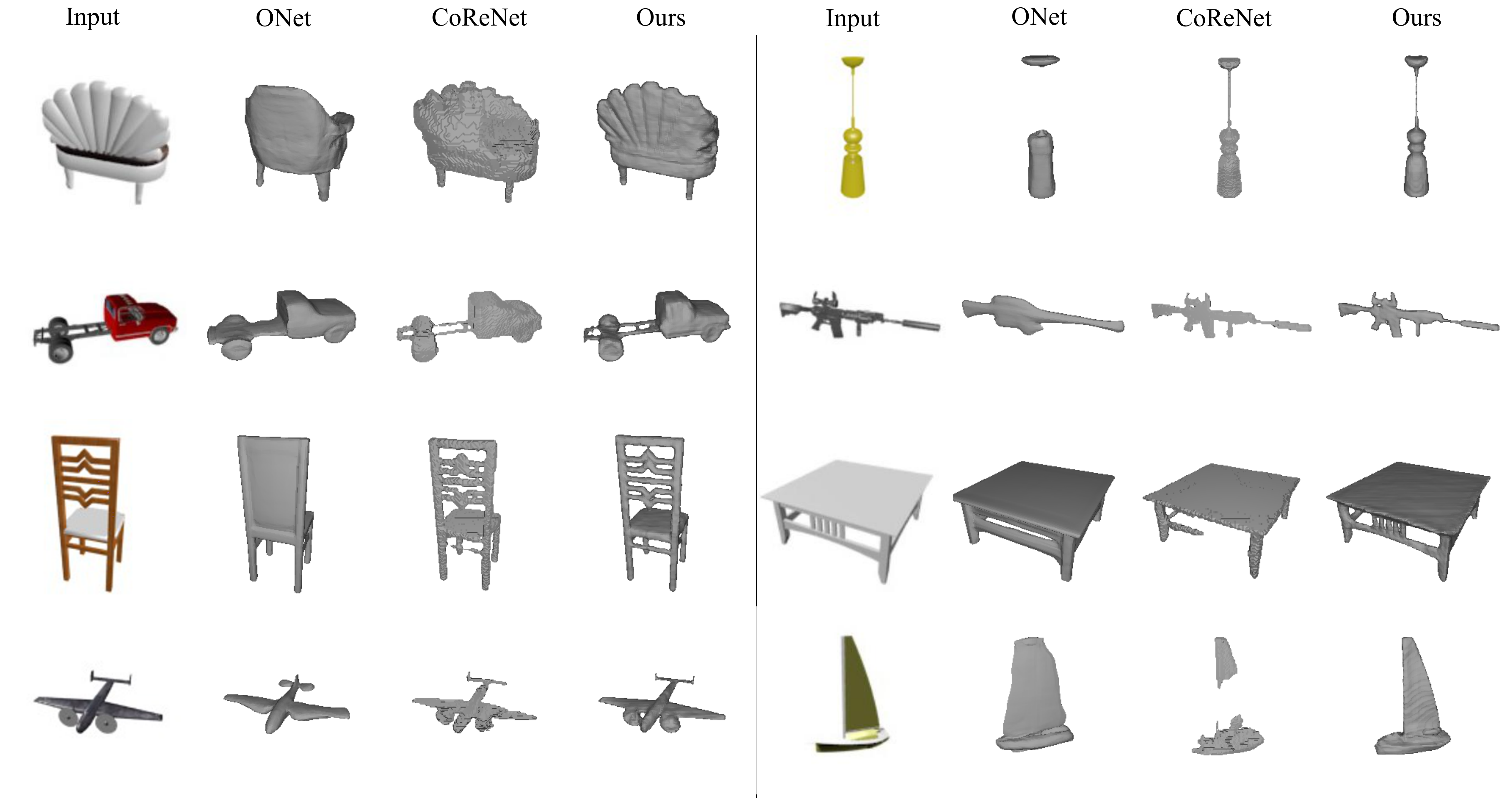}
    \caption{\textbf{Qualitative Reconstruction Results on ShapeNet.} Note that \methodname is able to reconstruct fine details of object shapes, such as the complicated pattern on the chair back.}
    \vspace{-0.4cm}
    \label{fig:vis_seen}
\end{figure}

\subsection{Inference}\label{sec:method_inference}
During inference, as we do not have access to camera-object distance, nor camera intrinsics, the scale calibration factor can be set to a random value in the range of the scaling factor used during training. 
As a result, our method recovers correct shapes in an unknown scale due to the inherent scale/depth ambiguity in the single-view 3D reconstruction task.

Before extracting meshes using the Marching Cubes algorithm \cite{lorensen1987marching}, we apply an additional re-sampling process on the \methodname occupancy predictions, which is in the shape of a camera frustum due to the nature of our ray-based method, to get occupancies in a regular 3D grid. This re-sampling process can be simply done by trilinear interpolation.

\begin{figure}[]
    \centering
    \includegraphics[width=0.95\textwidth]{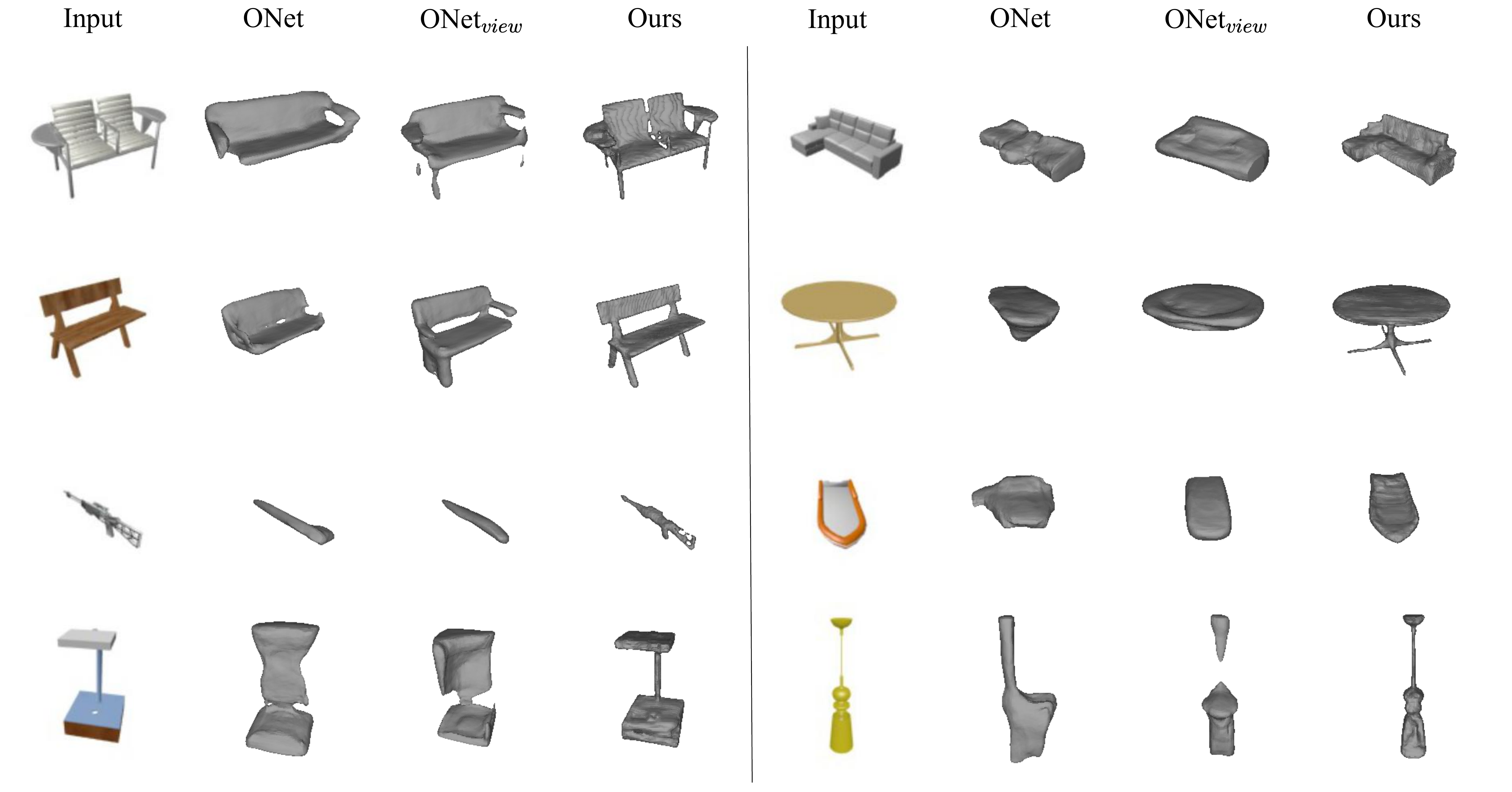}
    \caption{\textbf{Qualitative Reconstruction Results on Unseen Categories.} We show \methodname can generalise better than our baselines on unseen categories, in which object shapes are significantly different from the training classes (airplane, car, and chair).}
    \vspace{-0.4cm}
    \label{fig:vis_unseen}
\end{figure}

\section{Experiments} \label{sec:exp} 
We conduct experiments to evaluate our method on two benchmarks, showing our method achieves state-of-the-art performance while being much faster during inference. This section is organised as follows, \cref{sec:setup} reports the setup of our experiments. We compare our reconstruction quality with baselines in \cref{sec:acc}. Memory and time consumption are reported in \cref{sec:memo}. Lastly, we conduct an ablation study in \cref{sec:ablation}. 

\vspace{-0.15cm}
\subsection{Setup} \label{sec:setup}
\boldstart{Dataset.}
We evaluate \methodname on the ShapeNet dataset~\cite{chang2015shapenet}, which is a large-scale dataset containing 3D synthetic object meshes with category labels. To make fair comparisons with prior works, we follow the evaluation protocol in ONet \cite{mescheder2019occupancy}, taking a pre-processed subset from 3D-R2N2 \cite{choy20163d}, which includes image renderings from 13 classes in ShapeNetCore v1 \cite{chang2015shapenet} and adopting the same train/test splits. 

\boldstart{Metrics.}
Following ONet \cite{mescheder2019occupancy}, we adopt the volumetric IoU, the Chamfer-L1 distance, and the normal consistency score (NC) as the evaluation metrics. The volumetric IoU is evaluated on 100k randomly sampled points from the bounding volume. Additionally, we compute IoU on voxelised meshes in order to compare with some other baseline models. Methods with different IoU evaluation methods are separately listed in \cref{tab:seen_cat}. Excepting from \cref{tab:seen_cat}, our paper follows ONet's IoU evaluation method by default.

\boldstart{Implementation Details.}
\emph{Network initialisation}:
Apart from the image encoder $\mathcal{H}$, which is fine tuned from an ImageNet\cite{deng2009imagenet}-pretrained ResNet-18 backbone\cite{he2016resnet} with a random initialised fully-connected layer head, the local context mixer $\mathcal{J}$ and the occupancy decoder $\mathcal{D}$ are both trained from scratch.
\emph{Training details}:
During training, we randomly sample 1024 pixels on an input image and predict occupancies for 128 points that are equally spaced between a defined depth range on the rays.
We use a batch size of 64, with an Adam\cite{kingma2015adam} optimiser and a learning rate of $10^{-4}$. 
\emph{Network architecture}: A detailed description of network architectures is available in the supplementary material.

\vspace{-0.15cm}
\subsection{Reconstruction Accuracy}
\label{sec:acc}
\boldstart{Standard Reconstruction.}
We evaluate the performance of \methodname on the standard 3D reconstruction task, training/testing on all 13 categories. We compare our model with prior works including ONet~\cite{mescheder2019occupancy}, AtlasNet~\cite{groueix2018atlasnet}, CoReNet~\cite{popov2020corenet}, DmifNet~\cite{li2021dmifnet}, 3D43D~\cite{bautista2021generalization}, DISN~\cite{xu2019disn} and Ladybird~\cite{xu2020ladybird}.
Note that ONet, AtlasNet, DmifNet and DISN are object-centred models and the others are viewer-centred. DISN and Ladybird have different object scale from ours, hence we only evaluate IoU, which is scale-invariant.
\cref{tab:seen_cat} reports that our model achieves state-of-the-art 3D reconstruction accuracy on the ShapeNet benchmark.

\begin{table}[h]
\centering
\resizebox{\linewidth}{!}{%
\begin{tabular}{lccccccccccccc|c} 
\toprule
\textbf{IoU~}$\uparrow$          & airplane       & bench          & cabinet        & car            & chair          & display        & lamp           & loudspeaker    & rifle          & sofa           & table          & telephone      & vessel         & mean            \\ 
\hline
ONet                             & 0.591          & 0.492          & 0.750          & 0.746          & 0.530          & 0.518          & 0.400          & 0.677          & 0.480          & 0.693          & 0.542          & 0.746          & 0.547          & 0.593           \\
DmifNet                          & \textbf{0.603} & 0.512          & 0.753          & \textbf{0.758} & 0.542          & 0.560          & 0.416          & 0.675          & 0.493          & 0.701          & 0.550          & 0.750          & 0.574          & 0.607           \\
CoReNet$^\dagger$                                & 0.543          & 0.431          & 0.715          & 0.718          & 0.512          & 0.557          & 0.430          & 0.686          & 0.589          & 0.686          & 0.527          & 0.714          & 0.562          & 0.590           \\
3D43D                            & 0.571          & 0.502          & 0.761          & 0.741          & 0.564          & 0.548          & 0.453          & \textbf{0.729} & 0.529          & 0.718          & 0.574          & 0.740          & 0.588          & 0.621           \\
Ours                             & 0.574          & \textbf{0.521} & \textbf{0.763} & 0.755          & \textbf{0.567} & \textbf{0.601} & \textbf{0.467} & 0.726          & \textbf{0.594} & \textbf{0.726} & \textbf{0.578} & \textbf{0.757} & \textbf{0.598} & \textbf{0.633}  \\ 
\hline
DISN     & 0.617           & 0.542           & 0.531           & 0.770           & 0.549           & 0.577           & 0.397           & 0.559           & 0.680           & 0.671           & 0.489           & 0.736           & 0.602           & 0.594            \\
Ladybird & \textbf{0.666 } & 0.603           & 0.564           & 0.802           & \textbf{0.647 } & 0.637           & 0.485           & 0.615           & \textbf{0.719 } & 0.735           & 0.581           & \textbf{0.781 } & \textbf{0.688 } & 0.656            \\
Ours     & 0.630           & \textbf{0.630 } & \textbf{0.806 } & \textbf{0.815 } & 0.631           & \textbf{0.652 } & \textbf{0.549 } & \textbf{0.768 } & 0.662           & \textbf{0.768 } & \textbf{0.654 } & 0.773           & 0.658           & \textbf{0.692 }  \\
\hline
\textbf{Chamfer-L1~}$\downarrow$ & airplane       & bench          & cabinet        & car            & chair          & display        & lamp           & loudspeaker    & rifle          & sofa           & table          & telephone      & vessel         & mean            \\ 
\hline
AtlasNet$^*$                         & 0.104          & 0.138          & 0.175          & 0.141          & 0.209          & 0.198          & 0.305          & 0.245          & 0.115          & 0.177          & 0.190          & 0.128          & \textbf{0.151}          & 0.175           \\
ONet                             & 0.134          & 0.150          & 0.153          & 0.149          & 0.206          & 0.258          & 0.368          & 0.266          & 0.143          & 0.181          & 0.182          & 0.127          & 0.201          & 0.194           \\
DmifNet                          & 0.131          & 0.141          & 0.149          & 0.142          & 0.203          & 0.220          & 0.351          & 0.263          & 0.135          & 0.181          & 0.173          & 0.124          & 0.189          & 0.185           \\
CoReNet$^\dagger$  & 0.135 & 0.157 & 0.181 & 0.173 & 0.216 & 0.213 & 0.271 & 0.263 & \textbf{0.091} & 0.176 & 0.170 & 0.128 & 0.216 & 0.184  \\
3D43D                            & \textbf{0.096} & \textbf{0.112} & \textbf{0.119} & \textbf{0.122} & 0.193          & \textbf{0.166} & 0.561          & 0.229          & 0.248          & \textbf{0.125} & 0.146          & \textbf{0.107} & 0.175          & 0.184           \\
Ours                             & 0.121          & 0.124          & 0.140          & 0.145          & \textbf{0.178} & 0.183          & \textbf{0.238} & \textbf{0.204} & 0.094 & 0.151          & \textbf{0.140} & 0.109          & 0.163 & \textbf{0.153}  \\ 
\hline
\textbf{NC~}$\uparrow$            & airplane       & bench          & cabinet        & car            & chair          & display        & lamp           & loudspeaker    & rifle          & sofa           & table          & telephone      & vessel         & mean            \\ 
\hline
AtlasNet$^*$                         & 0.836          & 0.779          & 0.850          & 0.836          & 0.791          & 0.858          & 0.694          & 0.825          & 0.725          & 0.840          & 0.832          & 0.923          & 0.756          & 0.811           \\
ONet                             & 0.845          & 0.814          & 0.884          & 0.852          & 0.829          & 0.857          & 0.751          & 0.842          & 0.783          & 0.867          & 0.860          & \textbf{0.939} & 0.797          & 0.840           \\
DmifNet                          & \textbf{0.853} & 0.821          & 0.885          & \textbf{0.857} & 0.835          & 0.872          & 0.758          & 0.847          & 0.781          & 0.873          & 0.868          & 0.936          & 0.808          & 0.846           \\
CoReNet$^\dagger$  & 0.752 & 0.742 & 0.814 & 0.771 & 0.764 & 0.815 & 0.713 & 0.786 & 0.762 & 0.801 & 0.790 & 0.870 & 0.722 & 0.777  \\
3D43D                            & 0.825          & 0.809          & \textbf{0.886} & 0.844          & 0.832          & \textbf{0.883} & 0.766          & \textbf{0.868} & 0.798          & 0.875          & 0.864          & 0.935          & 0.799          & 0.845           \\
Ours                             & 0.830          & \textbf{0.828} & 0.881          & 0.849          & \textbf{0.844} & \textbf{0.883} & \textbf{0.780} & 0.866          & \textbf{0.837} & \textbf{0.881} & \textbf{0.873} & 0.929          & \textbf{0.815} & \textbf{0.854}  \\
\bottomrule
\end{tabular}}
\caption{\textbf{Reconstruction Accuracy on ShapeNet.} 
We show \methodname achieves state-of-the-art reconstruction accuracy in all three evaluation metrics. For IoU evaluation, the first group (5 rows) follows ONet's evaluation pipeline, the second group (3 rows) follows DISN's pipeline.
The notation $^*$ denotes results taken from \cite{mescheder2019occupancy} and $^\dagger$ denotes results we train their public code and evaluate in ONet's protocol. All other results are taken from corresponding original publications.
}
\vspace{-0.3cm}
\label{tab:seen_cat}
\end{table}

\boldstart{Reconstruction on Unseen Categories.}
To evaluate the generalisation capability, we train our model on 3 classes (airplane, car, and chair) and test on the remaining 10 classes. To make a fair comparison, we additionally train an ONet model with the viewer-centred setup, named ONet$_{view}$.
\cref{tab:unseen_cat} reports that our model performs better than all baselines in this setup. 

\begin{table}[h]
\centering
\resizebox{\linewidth}{!}{%
\begin{tabular}{lcccccccccc|c} 
\toprule
\textbf{IoU $\uparrow$}        & bench          & cabinet        & display        & lamp           & loudspeaker    & rifle          & sofa           & table          & telephone      & vessel         & mean            \\ 
\hline
ONet                & 0.251          & 0.400          & 0.165          & 0.147          & 0.394          & 0.172          & 0.505          & 0.222          & 0.147          & 0.374          & 0.278           \\
ONet$_{view}$           & 0.241          & 0.442          & 0.204          & 0.135          & 0.449          & 0.104          & 0.524          & 0.264          & 0.239          & 0.302          & 0.290           \\
3D43D               & 0.302          & 0.502          & 0.243          & 0.223          & 0.507          & 0.236          & 0.559          & 0.313          & 0.271          & \textbf{0.401} & 0.356           \\
Ours                & \textbf{0.334} & \textbf{0.517} & \textbf{0.256} & \textbf{0.246} & \textbf{0.538} & \textbf{0.253} & \textbf{0.583} & \textbf{0.350} & \textbf{0.277} & 0.394          & \textbf{0.375}  \\ 
\hline
\textbf{Chamfer-L1 $\downarrow$} & bench          & cabinet        & display        & lamp           & loudspeaker    & rifle          & sofa           & table          & telephone      & vessel         & mean            \\ 
\hline
ONet                & 0.515          & 0.504          & 1.306          & 1.775          & 0.635          & 0.913          & 0.412          & 0.629          & 0.659          & 0.392          & 0.774           \\
ONet$_{view}$           & 0.443          & 0.418          & 0.840          & 2.976          & 0.496          & 1.188          & 0.343          & 0.470          & 0.532          & 0.469          & 0.818           \\
3D43D               & 0.357          & 0.529          & 1.389          & 1.997          & 0.744          & 0.707          & 0.421          & 0.583          & 0.996          & 0.521          & 0.824           \\
Ours                & \textbf{0.228} & \textbf{0.351} & \textbf{0.592} & \textbf{0.485} & \textbf{0.389} & \textbf{0.307} & \textbf{0.266} & \textbf{0.299} & \textbf{0.453} & \textbf{0.304} & \textbf{0.367}  \\ 
\hline
\textbf{NC $\uparrow$}         & bench          & cabinet        & display        & lamp           & loudspeaker    & rifle          & sofa           & table          & telephone      & vessel         & mean            \\ 
\hline
ONet                & 0.694          & 0.697          & 0.527          & 0.563          & 0.693          & 0.583          & 0.779          & 0.701          & 0.605          & 0.694          & 0.654           \\
ONet$_{view}$           & 0.717          & \textbf{0.762} & 0.600          & 0.508          & 0.740          & 0.505          & 0.796          & 0.740          & 0.673          & 0.679          & 0.672           \\
3D43D               & 0.706          & 0.759          & 0.638          & 0.618          & 0.749          & 0.588          & 0.784          & 0.731          & \textbf{0.700} & 0.690          & 0.696           \\
Ours                & \textbf{0.736} & \textbf{0.762} & \textbf{0.640} & \textbf{0.642} & \textbf{0.762} & \textbf{0.609} & \textbf{0.806} & \textbf{0.764} & 0.670          & \textbf{0.703} & \textbf{0.709}  \\
\bottomrule
\end{tabular}
}
\caption{\textbf{Reconstruction Accuracy on Unseen Categories.} We report the performance of different models on 10 unseen categories after trained on 3 categories. Our method outperforms all prior works in almost all categories.
}
\label{tab:unseen_cat}
\end{table}

\vspace{-0.3cm}
\subsection{Memory and Inference Time}
\label{sec:memo}
As shown in \cref{tab:mem_time}, 
\methodname is over 20$\times$ faster than ONet even if the efficient Multi-resolution IsoSurface Extraction is used in ONet while maintaining a similar memory footprint.
It is worth noting that 3D-R2N2~\cite{choy20163d} uses a 3D auto-encoder architecture to predict on the discretised voxel grid. Although it is 4$\times$ faster than \methodname, it produces inferior reconstruction accuracy and requires a much larger memory size with 20$\times$ parameters.
Since SeqXY2SeqZ~\cite{han2020seqxy2seqz} has not released their code, we can only compare with them on complexity and accuracy. 
The inference time complexity of SeqXY2SeqZ is $O(N^2) \times l_{RNN}$, where $l_{RNN}$ denotes the number of the recurrent steps.
While being more efficient, we outperform SeqXY2SeqZ by 0.152 measured in IoU at $32^3$ resolution (SeqXY2SeqZ reports its IoU at this resolution).

\begin{table}[h]
\centering
\small
\begin{tabular}{lcccc} 
\toprule
Methods               & ONet  & 3D-R2N2 & CoReNet & Ours   \\ 
\hline
\# Params (M)          & 13.4 & 282.0  & 36.1   & 14.1  \\
Inference Memory (MB) & 1449  & 5113    & 2506    & 1559   \\
Inference time (ms)   & 333.1 & 3.4    & 37.8    & 13.8   \\
\bottomrule
\end{tabular}
\caption{\textbf{Comparisons of Number of Parameters, Inference Memory and Inference Time at $128^3$ Resolution}. The inference time measures the network forward time only and does not include any post-processing time.}
\vspace{-0.3cm}
\label{tab:mem_time}
\end{table}

\vspace{-0.3cm}
\subsection{Ablation Study}
\label{sec:ablation}
We analyse the effectiveness of the scale calibration factor $s$, the global and local image features as well as the architecture design.

\begin{figure}[h]
    \centering
    \includegraphics[width=0.95\textwidth]{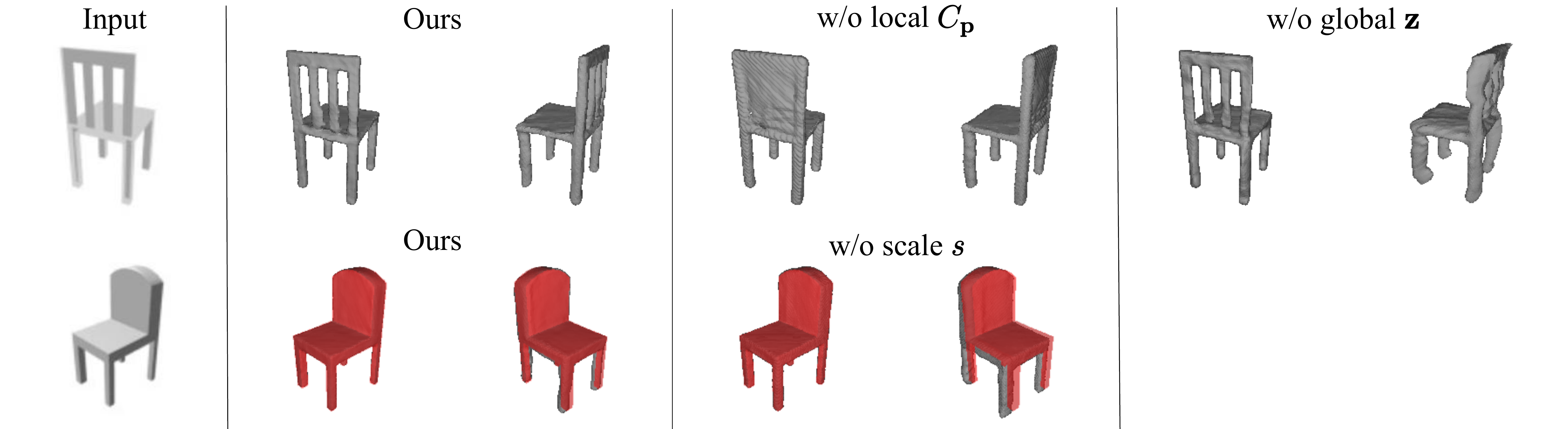}
    \caption{\textbf{Visualisation for Ablation Study.} Without the local feature, details of the object can not be recovered. Without the global feature, artifacts occur in other views. Without the scale factor, the scale of prediction (grey) can be different from ground truth (red). }
    \vspace{-0.4cm}
    \label{fig:vis_ablations}
\end{figure}

\boldstart{Effect of the Scale Calibration Factor.}
Training with the scale calibration factor can lead to better reconstruction accuracy as reported in \cref{tab:ablation}. 
This shows that decoupling scale from shape learning not only enables us to train with varying scales of ground-truth shapes but also helps in network convergence. As shown in \cref{fig:vis_ablations}, the scale factor helps to locate the object and reconstruct the object with correct scale.

\boldstart{Effect of Global and Local Image Features.}
Both global and local image features are important for reconstruction accuracy as suggested by the performance gap between w/ and w/o features in \cref{tab:ablation}. Visualisation in \cref{fig:vis_ablations} shows that local feature enables more details to be reconstructed, and global feature gives prior knowledge about object shape.

\boldstart{Effect of Architecture.}
We test a variation of our model by removing the Local Context Mixer and concatenating all inputs ($\mathbf{z}, s, C_\mathbf{p}, \mathbf{p}$) to feed into an Occupancy Decoder with MLPs. The evaluation mIoU for this variation is 0.610, showing our original design can achieve better performance.

\begin{table}[]
\centering
\begin{tabular}{lccc} 
\toprule
               & IoU $\uparrow$            & Chamfer-$L_1$ $\downarrow$  & NC $\uparrow$ \\ 
\hline
Ours & \textbf{0.633} & \textbf{0.153} & \textbf{0.854}     \\
w/o scale $s$   & 0.545           & 0.226           & 0.818               \\
w/o global $\mathbf{z}$  & 0.574           & 0.206           & 0.821               \\
w/o local $C_\mathbf{p}$   & 0.525           & 0.224           & 0.802               \\
\bottomrule
\end{tabular}
\caption{\textbf{Ablation Study.} We show the effectiveness of the scale calibration factor $s$, the global image feature $\mathbf{z}$ and the local image feature $C_\mathbf{p}$.}
\vspace{-0.4cm}
\label{tab:ablation}
\end{table}

\begin{wrapfigure}{r}{0.4\textwidth}
    \vspace{-1.5cm}
    \centering
    \includegraphics[width=0.4\textwidth]{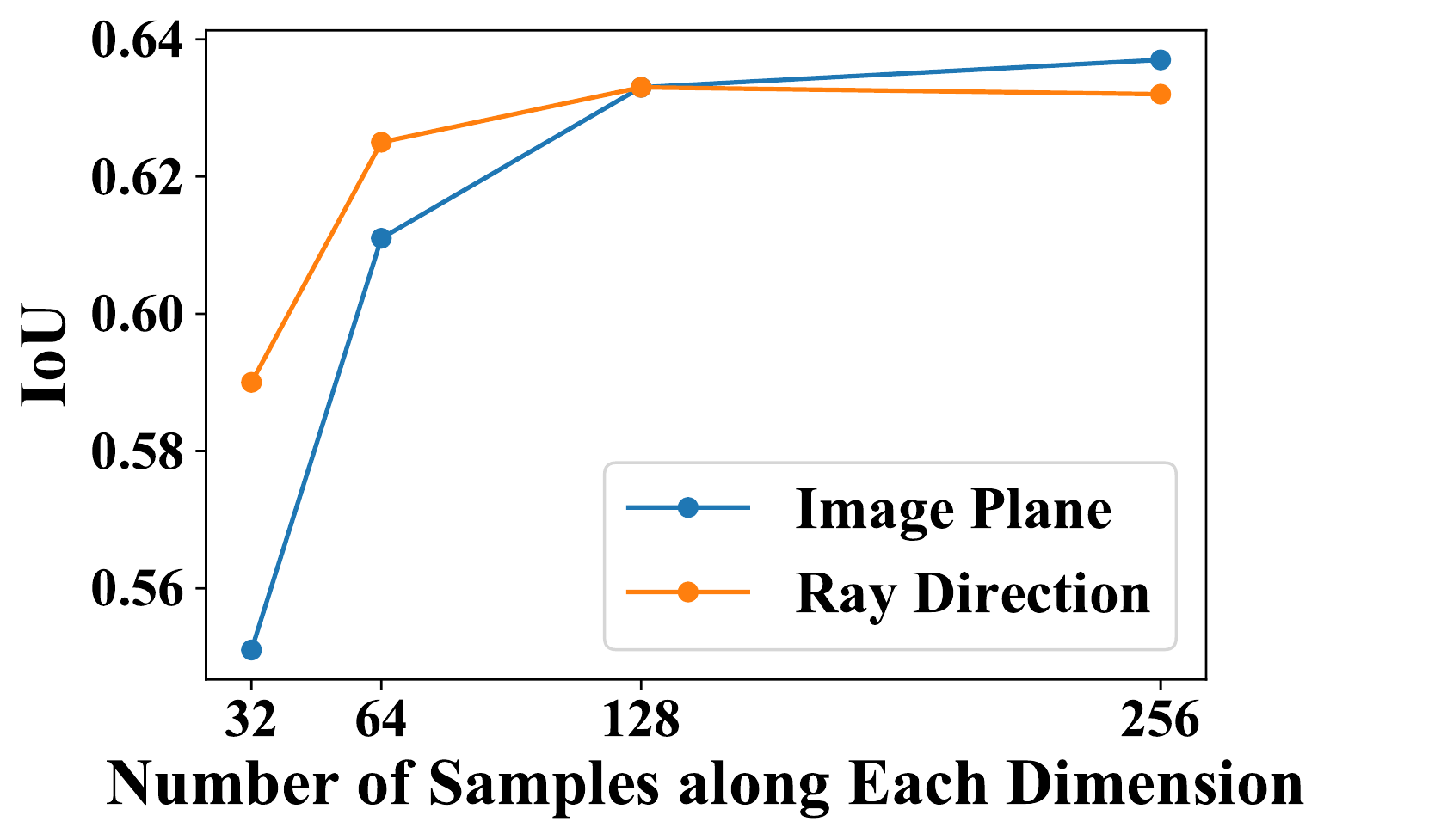}
    \caption{IoU of various No. of samples on a ray and in an image plane.}
    \vspace{-0.3cm}
    \label{fig:vis_plot}
\end{wrapfigure}

\vspace{-0.3cm}
\subsection{Discussion}
We study the effect of changing number of samples in the image-plane dimension $S_{plane}$ and along the ray direction $M$. 
\emph{Results:} \cref{fig:vis_plot} shows that although the reconstruction quality relates to both sampling dimensions, having $M>128$ along a ray cannot further improve reconstruction quality, whereas sampling more pixels in an image plane is still beneficial. 
\emph{Insights:} These results are not surprising as deducing occluded space is an ill-posed problem and can only be resolved from learning shape priors during training, which is bottlnecked by training data rather than $M=128$. On the other hand, since sampling more pixels in an image plane provides richer texture details, the reconstruction quality improves when $S_{plane}$ increases.
\emph{Setup:} 
To study the effect of different $S_{plane}$, we fix $M=128$ and test with $S_{plane}=32^2|64^2|128^2|256^2$. To study the effect of different $M$, we fix $S_{plane}=128^2$ and test with $M=32|64|128|256$.
\vspace{-0.3cm}
\section{Conclusion}
\label{sec:conclu}
We proposed \methodname, a deep neural network that reconstructs detailed object shapes efficiently given a single RGB image.
We demonstrate that predicting a series of occupancies along back-projected rays from both global and local features can significantly reduce the time complexity at inference while improving the reconstruction quality.
Experiments show that \methodname is able to achieve state-of-the-art performance for both seen and unseen categories in ShapeNet with a significant evaluation time reduction.
As our single-view object reconstruction results are encouraging, we believe an interesting direction in the future is to extend \methodname to tasks such as scene-level reconstructions or novel view synthesis.

\section*{Acknowledgements}
We thank Theo Costain for helpful discussions and comments. We thank Stefan Popov for providing the code for CoReNet and guidance on training. Wenjing Bian is supported by China Scholarship Council (CSC). 
\begin{appendices}
\title{Supplementary Material}
\section{Implementation Details}
The following sections include a detailed description of our model architecture, data preprocessing steps and evaluation procedure.

\begin{figure}[h]
    \centering
    \includegraphics[width=0.85\textwidth]{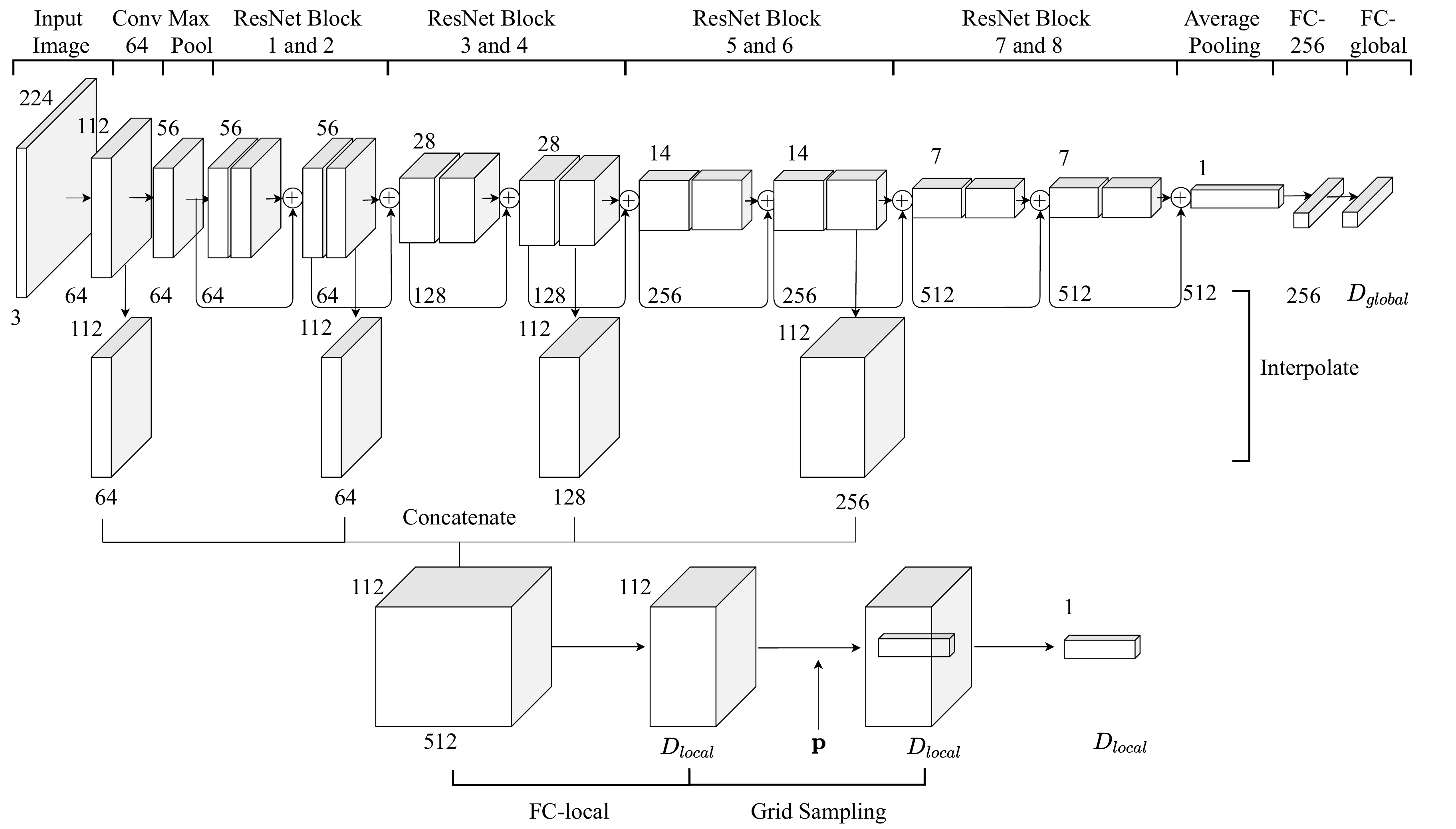}
    \caption{\textbf{Model Architecture for Encoder $\mathcal{H}$.} With a single image as input, the global latent code $\mathbf{z}$ is generated from a ResNet-18\cite{he2016resnet}. The local image feature is extracted from the position of 2D location $\mathbf{p}$ on the concatenated feature maps from 4 different encoder stages. Symbol $\oplus$ represents additive operation.}
    \vspace{-0.3cm}
    \label{fig:encoder}
\end{figure}
\begin{figure}[]
    \centering
    \includegraphics[width=0.85\textwidth]{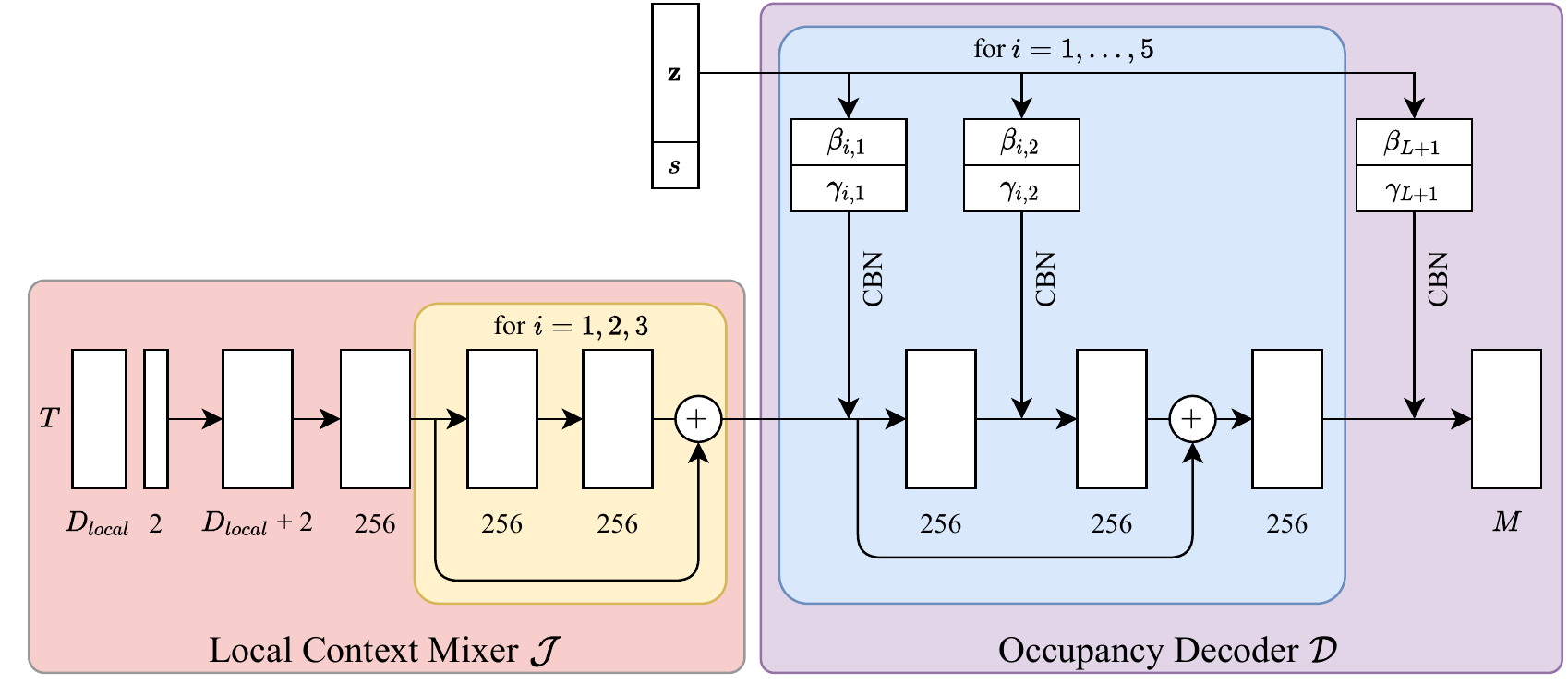}
    \caption{\textbf{Model Architecture for Local Context Mixer $\mathcal{J}$ and Occupancy Decoder $\mathcal{D}$.} Using the local feature and a 2D location $\mathbf{p}$ as input, we first generate a fused local context feature through Local Context Mixer $\mathcal{J}$. The Occupancy Decoder $\mathcal{D}$ uses Conditional Batch-Normalization (CBN) to condition the fused local feature on global latent code $\mathbf{z}$ and the scale calibration factor $s$, and predicts occupancy probabilities for $M$ points along the ray.}
    \vspace{-0.3cm}
    \label{fig:decoder}
\end{figure}

\boldstart{Encoder $\mathcal{H}$.}
The encoder is built on a ResNet-18 architecture\cite{he2016resnet} with an additional upsampling step to generate feature maps. The encoder is initialised with pretrained weights on ImageNet dataset\cite{deng2009imagenet} except the last fully connected layer. The global feature $\mathbf{z}$ is obtained from a fully connected layer with $D_{global}$ dimensions. The outputs from the $2^{nd}$, $4^{th}$ and $6^{th}$ ResNet Blocks, are upsampled to $112 \times 112$ using bilinear interpolation and concatenated, togenether with the the $112 \times 112$ feature maps output from the `Conv64' layer, to form 512 dimensional feature maps. The dimension is changed to $D_{local}$ with a fully connected layer. The local feature is then extracted from the corresponding position of 2D point $\mathbf{p}$ on the image. In practice, we choose $D_{global} = D_{local} =256$. 

\boldstart{Local Context Mixer $\mathcal{J}$.}
As shown in \cref{fig:decoder}, the Local Context Mixer $\mathcal{J}$ takes a batch of $T$ local features with $D_{local}$ dimensions and the corresponding 2D points as input. The local features and points are first concatenated and projected to 256 dimensions with a fully-connected layer. It then passes through 3 residual MLPs with ReLU activation before each fully-connected layer. The output local feature has $T$ batches with 256 dimensions. 

\boldstart{Occupancy Decoder $\mathcal{D}$.}
The Occupancy Decoder $\mathcal{D}$ follows the architecture of occupancy network\cite{mescheder2019occupancy}, with different inputs and output dimensions. The inputs are the global feature $(\mathbf{z}, s)$ and the local feature output from $\mathcal{J}$. The local feature first passes through 5 pre-activation ResNet-blocks. Each ResNet-block consist of 2 sub-blocks, where each sub-block applies Conditional Batch-Normalization (CBN) to the local feature followed by a ReLU activation function and a fully-connected layer. The output from the ResNet-block is added to the input local feature. After the 5 ResNet-blocks, the output passes through a last CBN layer and ReLU activation and a final fully-connected layer which produces the $M$ dimensional output, representing occupancy probability estimations for $M$ points along the ray. 
\subsection{Data Preprocessing}
We use the image renderings and train/test split of ShapeNet\cite{chang2015shapenet} as in 3D-R2N2\cite{choy20163d}. Following ONet\cite{mescheder2019occupancy}, the training set is subdivided into a training set and a validation set. 

In order to generate ground truth occupancies for rays from each view, we first create watertight meshes using the code provided by \cite{stutz2018learning}. With camera parameters for 3D-R2N2 image renderings, we place the camera at corresponding location of each view, and generate 5000 random rays passing through the image. We sample equally spaced points on each ray between defined distances $d_{min}$ and $d_{max}$. In practice, we choose $d_{min}=0.63$ and $d_{max}=2.16$, which guarantee all meshes are within this range. 

\subsection{Evaluation}
To make a fair comparison with previous approaches, we use normalised ground truth points and point clouds produced by ONet\cite{mescheder2019occupancy} for evaluation. With the scale calibration factor, our predicted mesh has the same scale as the raw unnormalised ShapeNet mesh. In order to make the predicted mesh in a consistent scale as the normalised ground truth, we use the scale factor between the raw ShapeNet mesh and the normalised ShapeNet mesh to scale our prediction before evaluation. The threshold parameter for converting occupancy probabilities into binary occupancy values is set to 0.2 during inference.

\section{Additional Experimental Results}
\subsection{Additional Qualitative Results on ShapeNet}
\begin{figure}[]
    \centering
    \includegraphics[width=0.95\textwidth]{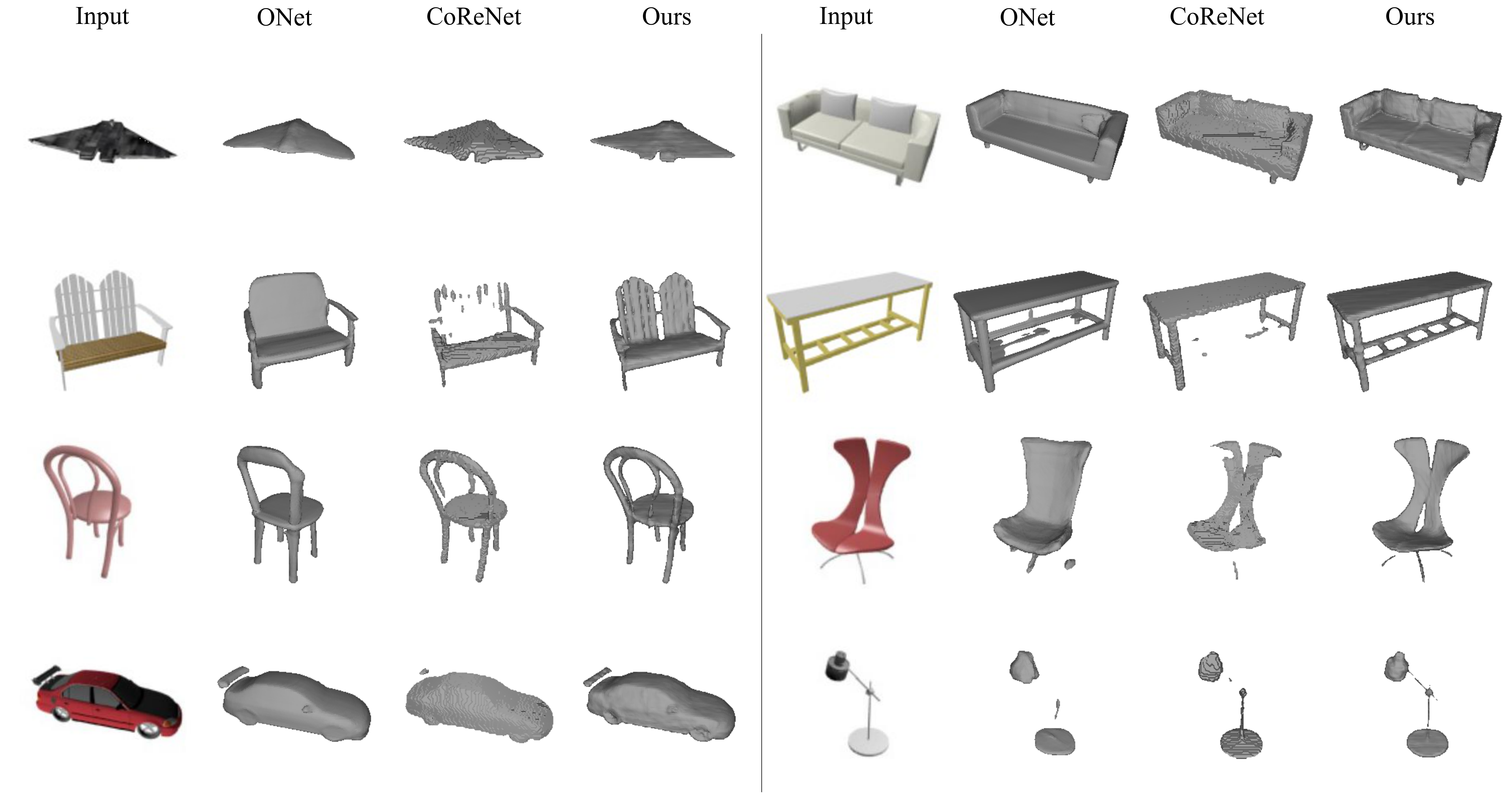}
    \caption{\textbf{Additional Qualitative Reconstruction Results on ShapeNet.}}
    \label{fig:vis_2}
\end{figure}

\begin{figure}[]
    \centering
    \includegraphics[width=0.95\textwidth]{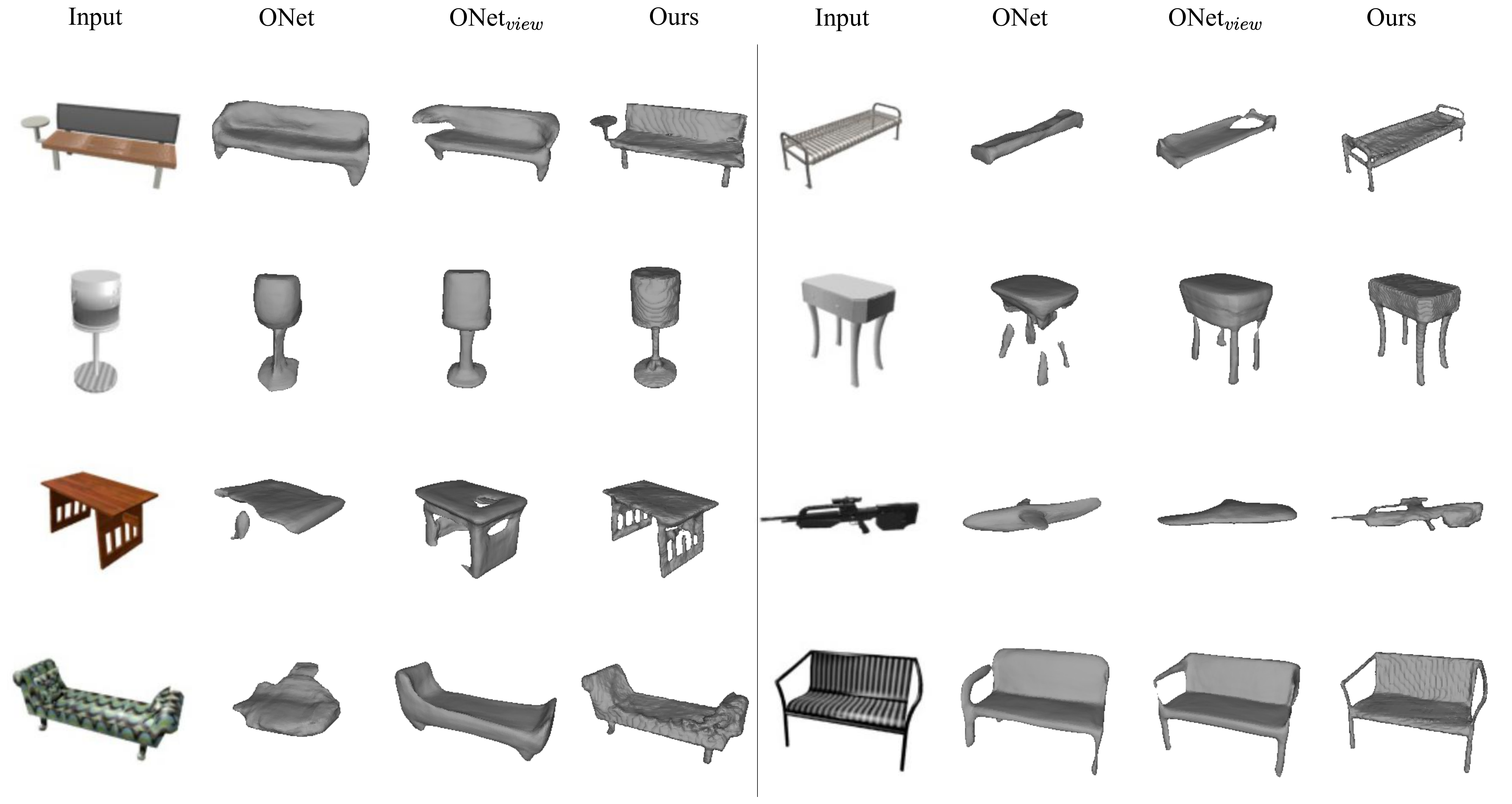}
    \caption{\textbf{Additional Qualitative Reconstruction Results on Unseen Categories.}}
    \label{fig:unseen_2}
\end{figure}

Additional qualitative results on ShapeNet are shown in \cref{fig:vis_2} and \cref{fig:unseen_2}, where \cref{fig:vis_2} is the standard reconstruction task results on categories seen during training, and \cref{fig:unseen_2} is the test results after trained on 3 different categories.
\subsection{Qualitative Results on Online Products and Pix3D Datasets}
With the model trained on 13 categories of synthesis ShapeNet\cite{chang2015shapenet} images, we made further tests on 2 additional datasets with real world images to validate the generalisation ability of the model.

\begin{figure}[h]
    \centering
    \includegraphics[width=0.85\textwidth]{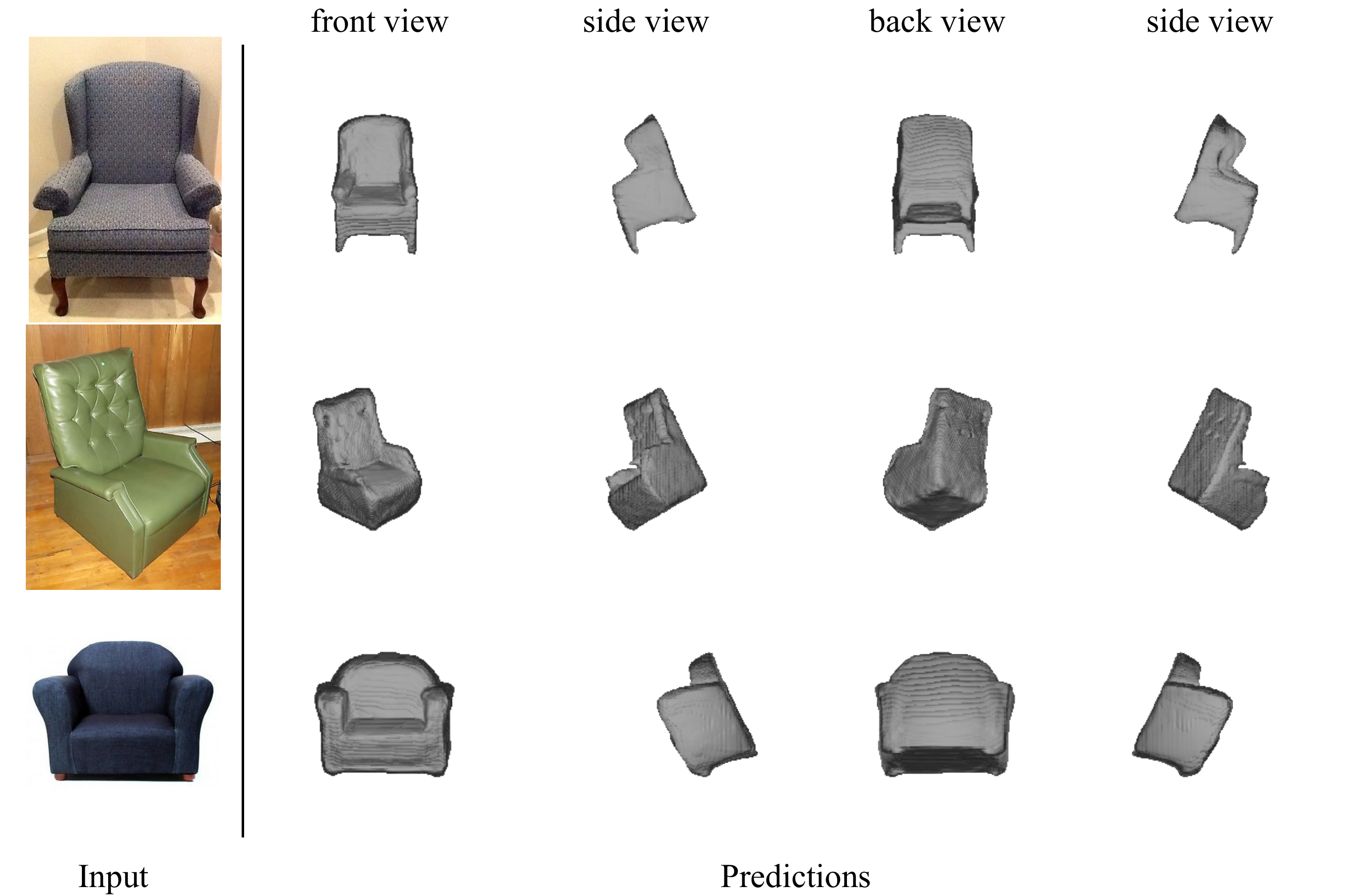}
    \caption{\textbf{Qualitative Reconstruction Results on Online Products dataset.}}
    \label{fig:online}
\end{figure}

\boldstart{Online Products Dataset.}
We use the chair category in Online Products dataset\cite{oh2016deep}.
As the training data has white background, We first feed the image into DeepLabV3\cite{chen2017deeplab} to generate a segmentation mask and changed the color of the image outside the mask to white. As there is no camera parameters available, the reconstruction shown in \cref{fig:online} is of correct proportion only.

\begin{figure}[h]
    \centering
    \includegraphics[width=0.85\textwidth]{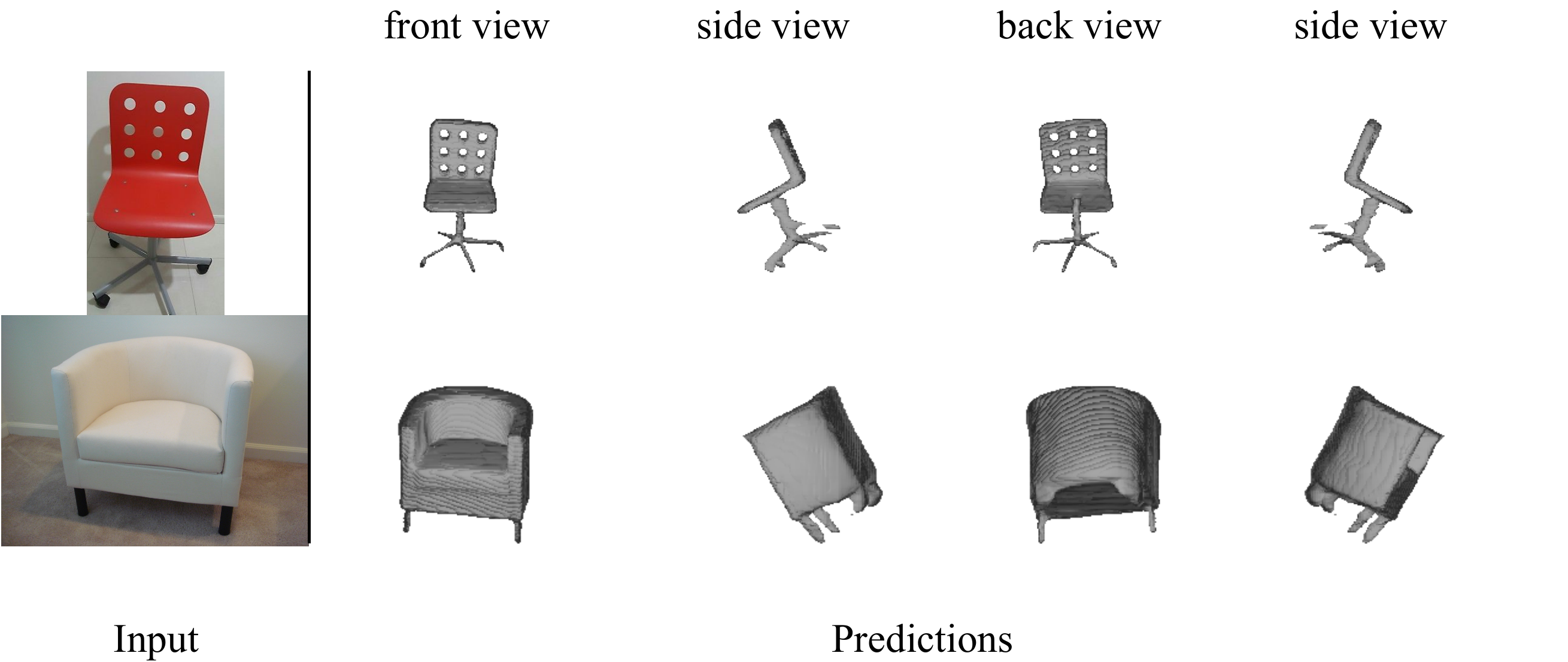}
    \caption{\textbf{Qualitative Reconstruction Results on Pix3D dataset.}}
    \label{fig:pix3d}
\end{figure}

\boldstart{Pix3D Dataset.} 
Similarly, we test our model on chair category of Pix3D dataset\cite{sun2018pix3d}, with the ground truth segmentation mask provided. Some results are shown in \cref{fig:pix3d}.

\vspace{-0.5cm}
\subsection{Limitations}
\begin{figure}[h]
    \centering
    \includegraphics[width=0.85\textwidth]{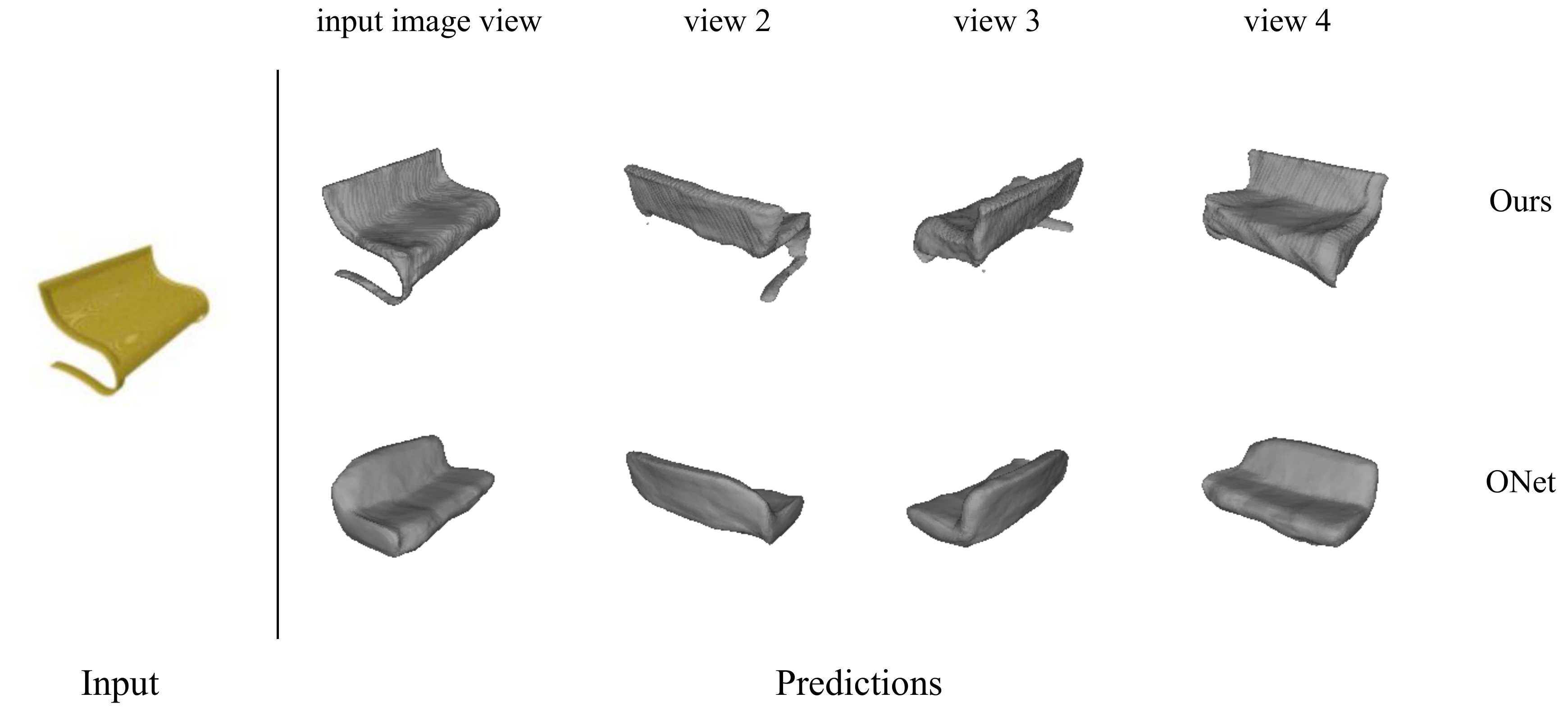}
    \caption{\textbf{A Failure Case for Novel Categories.}}
    \label{fig:failure1}
\end{figure}
As shown in \cref{fig:failure1}, for certain objects in unseen categories, the 3D object reconstructions are much more accurate in the image view than other views, as our model is able to predict shape for visible parts from image features but lacks of shape priors for invisible parts on those unseen categories.

\end{appendices}
\clearpage
\bibliography{egbib}
\end{document}